\documentclass[letterpaper]{article} 
\usepackage{aaai2027}  
\usepackage[hyphens]{url}  
\usepackage{graphicx} 
\usepackage{natbib}  
\usepackage{caption} 
\usepackage{booktabs}
\usepackage{url}
\usepackage{makecell}
\usepackage{multirow}

\usepackage{algorithm}
\usepackage{algorithmic}
\usepackage{amsmath}

\title{CADIR: A Cross-Backend Editable Intermediate Representation for Agentic CAD Generation}
\author{
    Yu Liu\equalcontrib,
    Jingzhe Ni\equalcontrib,
    Yiming Chen,  Junqi Huang, 
    Ruofeng Tong,
    Min Tang,
    Peng Du\thanks{Corresponding author.}
}

\affiliations{
    Zhejiang University, China\\
    dp@zju.edu.cn
}

\begin{document}
\nocopyright
\maketitle

\begin{abstract}
Large language models have made it possible to generate executable computer-aided design (CAD) programs from natural-language descriptions or images. However, existing methods represent modeling processes as backend-specific sequential scripts with implicit dependencies or as static geometry, making it difficult to simultaneously preserve construction history, stable topological references, and feature-level editability across different CAD systems. We present CADIR, an agent-friendly executable intermediate representation for CAD generation and cross-backend editing. Built on the OCCT geometry kernel via OCP, CADIR provides explicit, compositional modeling operations and fine-grained execution diagnostics. During program execution, CADIR records modeling operations, parameter dependencies, constraints, and topology selections in a construction graph. To enable reliable cross-backend reconstruction, we introduce Geometric Signature Matching, which identifies corresponding edges and faces despite parameter changes and backend differences, allowing adapters to reconstruct native editable feature histories in FreeCAD, SolidWorks, and Fusion 360. Building on this representation, we further propose a construction-graph retrieval method for text and image queries that supports both full-graph and subgraph retrieval, enabling agents to leverage complete models and modeling substructures. Extensive experiments demonstrate that CADIR achieves higher geometric fidelity and execution reliability than existing CAD representations, that construction-graph retrieval further improves model generation quality, and that cross-backend editing enables reliable model reconstruction and post-reconstruction editing across multiple CAD environments. 
The code is publicly available at \textcolor{blue}{\url{https://github.com/NiJingzhe/SimpleCADAPI}}.
\end{abstract}

\section{Introduction}

Computer-Aided Design (CAD) underpins modern engineering and manufacturing. CAD models are not defined solely by final geometry, but are built through sketching, extrusion, revolution, chamfering, filleting, Boolean operations, patterning, and constraints. These sequences encode construction history and design intent, enabling parameter modification, feature replacement, and dependency adjustment. Advances in large language models (LLMs) and multimodal generative models have made natural-language-driven CAD generation a promising way to lower the modeling barrier: users describe a part, and the model generates executable CAD programs or 3D geometry, advancing human--machine collaborative design~\cite{wu2026AICADsurvey}.

\begin{figure*}[t]
\centering
\includegraphics[width=0.95\textwidth]{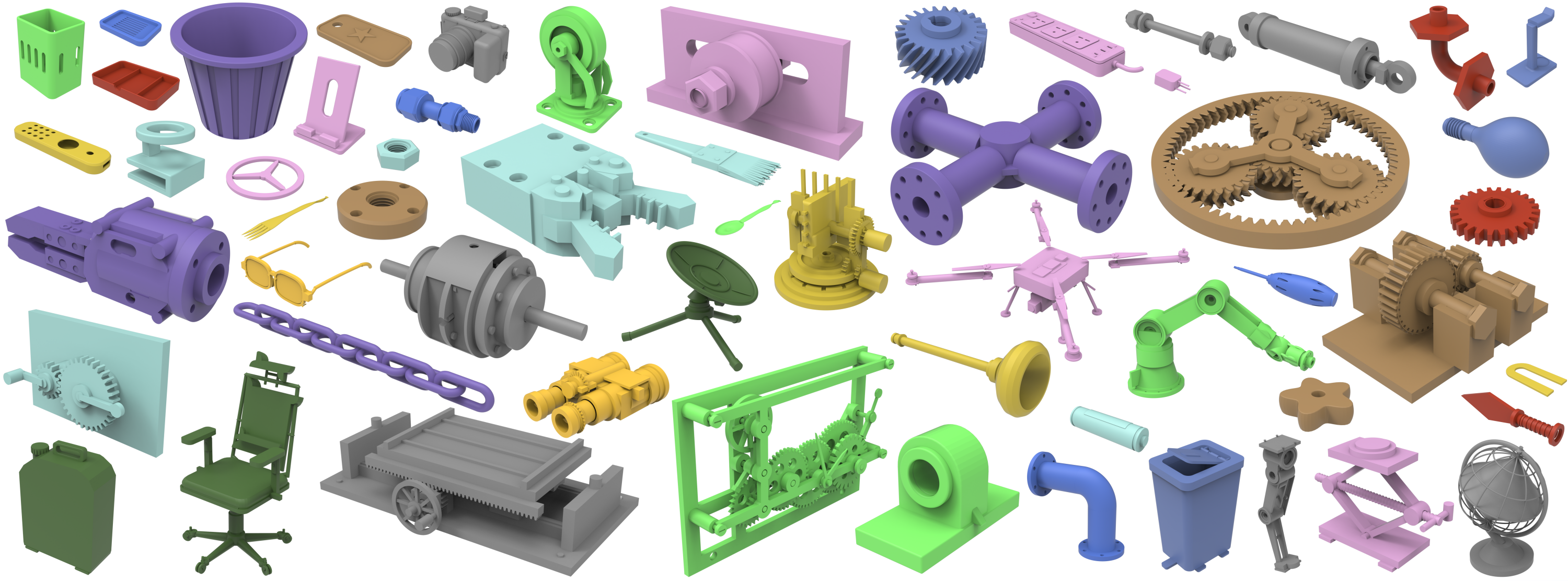}
\caption{Examples of CAD models generated by CADIR Agent across diverse mechanical and product-design tasks. The results include individual parts and assemblies involving Boolean operations, fillets, repeated features, and complex local structures, demonstrating the system's ability to generate geometrically diverse CAD models.}
\label{fig:generation_capability}
\end{figure*}

In recent years, CAD generation has witnessed multi-faceted progress. B-Rep or mesh methods, such as BrepGen~\cite{xu2024Brepgen} and AutoBrep~\cite{xu2025AutoBrep}, produce static geometry without construction history, preventing parametric and feature-level editing. Sequence-based methods, such as DeepCAD~\cite{wu2021Deepcad} and Text2CAD~\cite{khan2024Text2cad}, retain partial modeling history but are largely limited to sketch--extrude sequences. CAD-code methods, such as FutureCAD~\cite{li2026FutureCAD} and Cad-coder~\cite{guan2026Cad-coder}, support broader operations but rely on backend-specific flat scripts lacking explicit dependencies, constraints, parameter bindings, and stable topological references. Agent-based frameworks, such as CAD-Assistant~\cite{mallis2025CAD-assistant} and ProCAD~\cite{yuan2026ProCAD}, introduce execution feedback and iterative repair. However, they remain tied to specific CAD environments and lack a structured representation for cross-backend reconstruction or for retrieving complete models and modeling substructures.

To address these limitations, we propose CADIR, an agent-friendly and executable CAD intermediate representation for cross-backend editing. Our main contributions are as follows:
\begin{itemize}
    \item We propose CADIR, an agent-friendly executable CAD interface built on OCP/OCCT. It exposes explicit compositional operations, semantic topology- and geometry-aware selection, assembly constraints, and structured diagnostics for reliable CAD construction and repair.

    \item We propose a cross-backend reconstruction method that leverages the construction graph produced during CADIR execution and Geometric Signature Matching (GSM) to establish topological correspondence across CAD systems. With adapters for FreeCAD, SolidWorks, and Fusion~360, the method reconstructs native editable feature histories in different CAD backends.

    \item We propose a construction-graph retrieval mechanism for text and image queries that supports both full-graph and subgraph retrieval, enabling agents to reuse complete models and modeling substructures for complex CAD generation.
\end{itemize}

\begin{figure*}[t]
\centering
\includegraphics[width=0.95\textwidth]{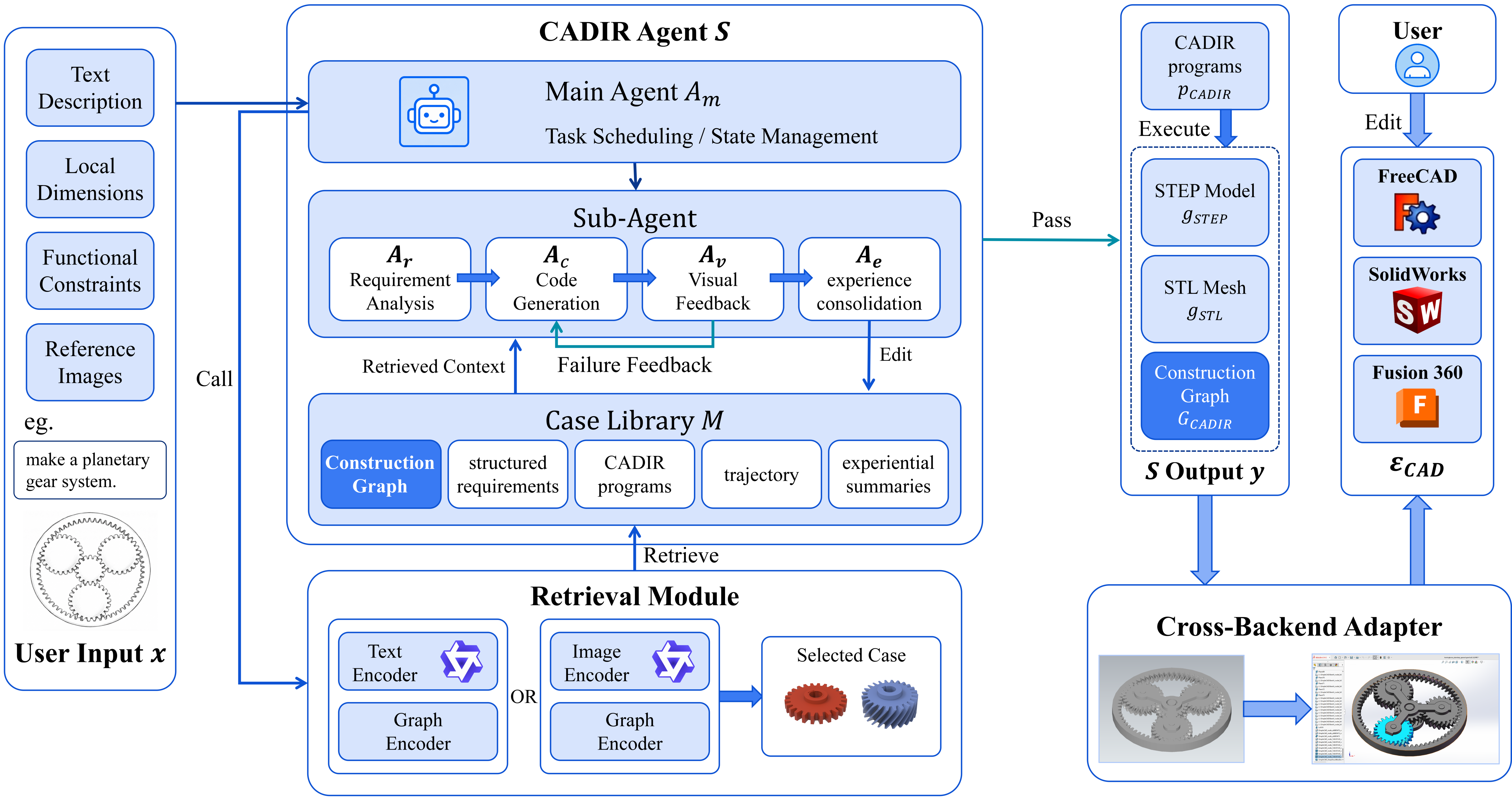}
\caption{Overview of the CADIR system. Given multi-modal inputs, the CADIR Agent generates CADIR programs and construction graphs, which are replayed by cross-backend adapters to reconstruct native editable models in familiar CAD environments such as FreeCAD, SolidWorks, and Fusion~360.}
\label{fig:cadir_agent_framework}
\end{figure*}

\section{Related Work}
This section introduces prior work relevant to our research: agent-based CAD generation, editable intermediate representations, and retrieval-augmented generation.

\subsection{Agent‑Based Modeling Script Generation}
CAD generation methods that primarily rely on offline data training (e.g., DeepCAD \cite{wu2021Deepcad}, Text2CAD \cite{khan2024Text2cad}, BrepDiff \cite{lee2025Brepdiff}) suffer from quality heavily dependent on training data distributions due to the lack of execution feedback and automatic correction. In contrast, agent‑based methods leverage closed‑loop self‑optimization and self‑verification mechanisms, enabling models to continuously improve through execution feedback. CAD‑Assistant \cite{mallis2025CAD-assistant} integrates geometric evaluation tools to enhance generation quality, yet its linear instruction structure exhibits limited adaptability in heterogeneous API environments. Seek‑CAD \cite{li2025Seek-CAD} employs a local LLM with visual chain‑of‑thought to achieve training‑free generation, offering efficiency advantages. However, its flat script output is difficult to accommodate incremental requirement changes through localized edits. From Idea to CAD \cite{ocker2025idea-CAD} and ProCAD \cite{yuan2026ProCAD} propose multi‑agent collaboration and proactive interactive clarification, but they have not established clear error‑recovery mechanisms to handle exception backtracking in long‑horizon planning. Overall, existing methods emphasize task decomposition and interaction, but provide limited support for heterogeneous CAD APIs and structured recovery from execution failures.

\subsection{Intermediate Representations for Editable CAD}

Existing modeling scripts are typically tied to a single modeling language, making cross-language translation and cross-engine editing difficult. To address this limitation, researchers have proposed intermediate representations that encode CAD modeling sequences, including sketching, extrusion, revolution, filleting, and patterning, into structured formats suitable for large language models. FlexCAD~\cite{zhang2025Flexcad} adopts layer-aware masking for fine-grained control, but its structured linear representation does not address generalization across geometry-kernel APIs. HistCAD~\cite{dong2025HistCAD} proposes a constraint-aware parametric history representation, yet it is not directly executable and lacks a general mechanism for cross-backend reconstruction. CAD-Llama~\cite{li2025CAD-Llama} converts command sequences into Structured Parametric CAD Code (SPCC), a code-like representation compatible with LLM inputs, but provides limited support for runtime dependency tracking and failure recovery. CADDesigner~\cite{fan2026CADDesigner} combines agent workflows with an explicit modeling language to support iterative CAD generation, but does not address cross-engine reconstruction and editing of modeling histories. ArtisanCAD~\cite{xu2026ArtisanCAD} introduces CAD-IR within the CATIA-MCP environment, but remains tied to the CATIA backend. Overall, existing representations provide limited support for cross-API execution, cross-backend reconstruction, and structured recovery from incremental edits.

\subsection{Retrieval-Augmented Generation}

Retrieval-augmented generation improves complex model generation by reusing historical construction experience. GenCAD~\cite{alam2024Gencad} validates the effectiveness of using images as query modalities for retrieving CAD knowledge through contrastive learning, yet its retrieval units are confined to complete scripts, making it difficult to locate reusable local constructions. MemRL~\cite{zhang2026Memrl} leverages episodic memory to enhance continual agent evolution, but focuses on memory storage and replay strategies, with insufficient exploration of structured indexing of operation dependencies. GeAR~\cite{shen2025GeAR} and GraphCodeAgent~\cite{li2025GraphCodeAgent} provide valuable references for structured matching via graph-enhanced retrieval. Collectively, these methods rely on visual features, complete scripts, or generic graph representations as retrieval units, which limits fine-grained retrieval of local construction subgraphs.

\section{CADIR System Overview}

In this section, we first formulate the task and then introduce the CADIR Agent framework. As shown in Figure~\ref{fig:cadir_agent_framework}, the framework maps user inputs to editable CAD models in target environments through CADIR Agent and backend adapters.

\subsection{Problem Formulation}

Existing text-to-CAD and image-to-CAD methods typically produce static geometry, backend-specific scripts, or modeling sequences tied to a single CAD environment. These outputs cannot simultaneously preserve construction history, support reconstruction in the user's preferred CAD system, and provide stable references for complex modeling and topology-selection operations. We therefore formulate CAD generation in two stages. Given natural-language or image input $x$, CADIR Agent first produces an executable, inspectable set of intermediate artifacts $\mathcal{Y}$. The adapter translates $\mathcal{Y}$ into an editable document in the target CAD environment:
\[
x \stackrel{\mathrm{Agent}}{\longrightarrow} \mathcal{Y}
\stackrel{\mathrm{Adapter}}{\longrightarrow} \mathcal{E}_{\mathrm{CAD}},
\]
where $\mathcal{E}_{\mathrm{CAD}}$ denotes an editable document in a target environment such as FreeCAD, SolidWorks, or Fusion 360. The intermediate artifact set is
\[
\mathcal{Y}=\{p_{\mathrm{CADIR}},g_{\mathrm{STEP}},g_{\mathrm{STL}},G_{\mathrm{CADIR}}\},
\]
where $p_{\mathrm{CADIR}}$ is the generated CADIR program, $g_{\mathrm{STEP}}$ and $g_{\mathrm{STL}}$ are the exported CAD geometry and mesh, respectively, and $G_{\mathrm{CADIR}}$ is the construction graph recorded from the program's execution trace and replayed by the adapter to reconstruct an editable model in the target environment.

\subsection{Agent Architecture}

CADIR Agent consists of five agents and a retrievable case library:
\[
\mathcal{S}=(\mathcal{A},M),\qquad
\mathcal{A}=\{A_m,A_r,A_c,A_v,A_e\}.
\]
Here, $A_m$ coordinates the overall workflow; $A_r$, $A_c$, $A_v$, and $A_e$ perform requirement analysis, code generation and repair, visual verification, and experience consolidation, respectively. The case library $M$ organizes validated trajectories using construction graphs as the primary retrieval index.

The main agent repeatedly coordinates retrieval, generation, execution, validation, and repair until the model is accepted, after which the validated trajectory is stored in $M$.

During benchmark evaluation, the case library is frozen, and no trajectory from the test set is added to $M$. Outside evaluation, validated trajectories may be consolidated into the library for continual improvement.

\section{Construction-Graph-Centered CADIR}

In this section, we introduce CADIR from three perspectives: the agent-friendly modeling API, the construction graph for cross-backend reconstruction, and construction-graph retrieval over full graphs and local subgraphs.

\subsection{Agent-Friendly Modeling API}

Python modeling interfaces such as CadQuery\cite{2026cadquery} have established the utility of programmatic CAD, but their abstractions are designed primarily for human programmers. Implicit workplanes and stateful method chaining can obscure the active modeling context for an LLM. Heterogeneous return types and unstable rules for selecting edges and faces further complicate diagnosis and repair. To address these limitations, we design CADIR with 115 public modeling operations exposed through explicit, compositional calls.

CADIR defines an executable modeling boundary directly on the OCP/OCCT geometry kernel. Direct kernel access supports expressive geometric and constraint parameters while returning precise analytic geometry, topology, diagnostic logs, and execution errors that help the agent localize failures. Each API call also records its runtime state, which is subsequently encoded in the construction graph. 

During the modeling process, selecting an edge or face by index alone is unreliable because Boolean operations and parameter changes may reorder or alter the topology.  CADIR therefore defines a Topology Selector (TS) as an explicit selection operation, represented by
$\xi=\langle\tau,\rho,\pi,\phi,o,k,c\rangle$.
Here, $\tau$ specifies the target topology type; $\rho$ binds the search scope to a construction-graph output; $\pi$ defines optional boundary traversal; $\phi$ combines semantic tags and geometric predicates; $o$ and $k$ control ordering and truncation; and $c$ specifies the expected cardinality. Its grammar is
\[
\begin{aligned}
\mathsf{TSel} ::= {}& \texttt{Select}(\tau,\rho)
[\texttt{Traverse}(\pi)][\texttt{Where}(\phi)]\\
&[\texttt{Order}(o)][\texttt{Take}(k)]
[\texttt{Card}(c)].
\end{aligned}
\]
CADIR evaluates the selector stages in sequence and reports an error when the source is missing, the topology type is invalid, the traversal is unsupported, or the selected entities violate the expected cardinality. Upon successful selection, CADIR records the selector as a TS operation node in the construction graph. The selected entities and their geometric signatures are stored in the node’s structured operation record for backend translation.

CADIR provides explicit assembly operations for fixed, prismatic, revolute, gear, timing-belt, and rack-and-pinion constraints. These constraints encode component placement and relative-motion relationships and are recorded in the construction graph. For assembly diagnosis, CADIR provides a static collision verifier that can inspect scoped component pairs and return structured collision reports.

\subsection{Cross-Backend Construction Graph}

Cross-backend CAD exchange typically transfers final boundary representations or reimplements modeling procedures using backend-specific APIs. Geometry exchange discards construction history and parameter dependencies, whereas script-level conversion couples replay to platform-specific operations and unstable topology indices. CADIR addresses both limitations by recording program execution as a backend-independent construction graph.

\begin{figure}[t]
\centering
\includegraphics[width=0.95\columnwidth]{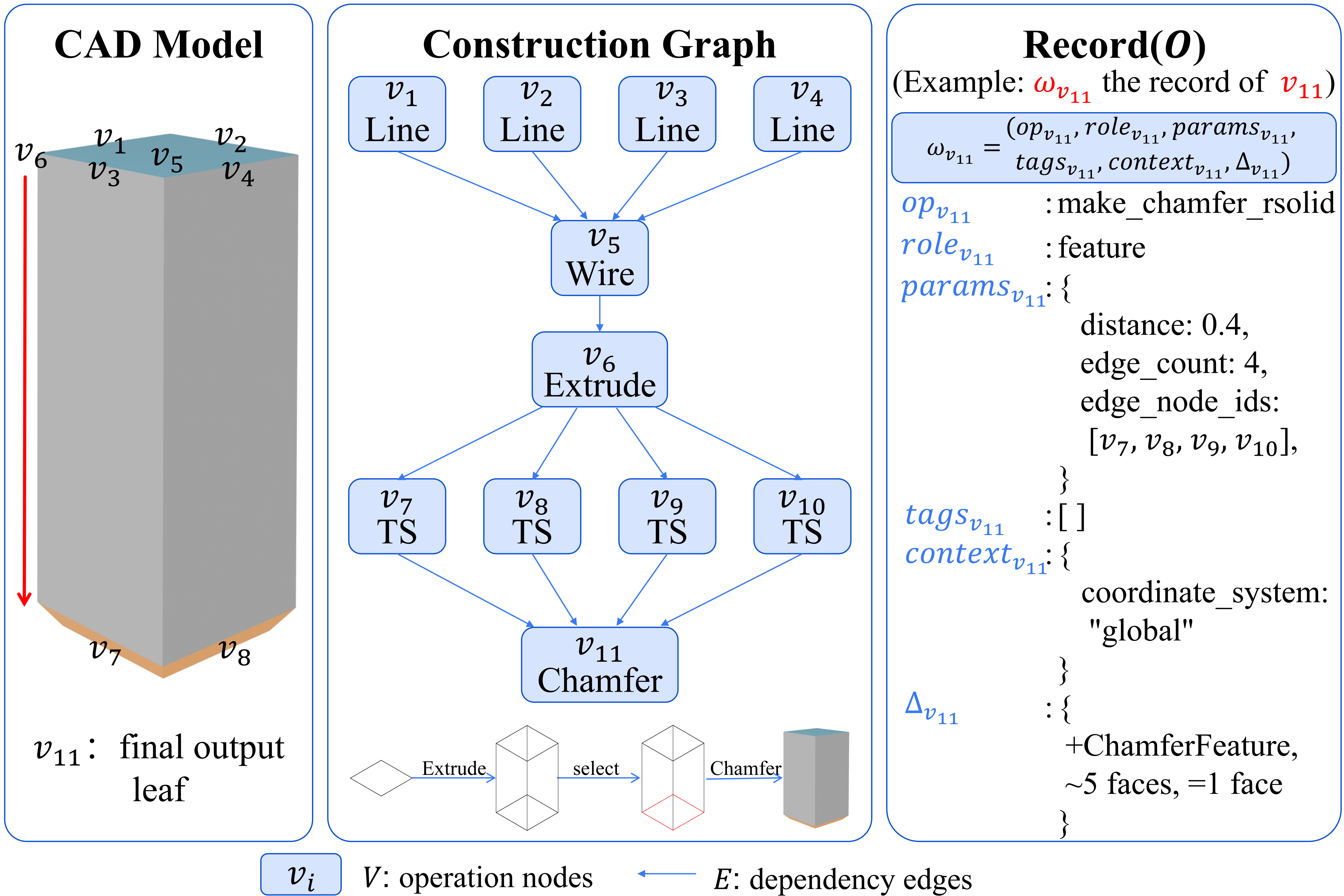}
\caption{Example of a CADIR Construction Graph. Nodes represent modeling operations, directed edges encode dependencies, and each node stores a structured operation record. Result nodes correspond to the final CAD output.}
\label{fig:feature_graph_definition}
\end{figure}

We denote the serialized construction graph $G_{\mathrm{CADIR}}$ by $G$ and define it as
\[
G=(V,E,\mathcal{O},\mathcal{R}),
\]
where $V=\{v_i\}_{i=1}^{n}$ is the set of canonical construction nodes, $E\subseteq V\times V$ is the set of operation-dependency edges, $\mathcal{O}=\{\omega_i\}_{i=1}^{n}$ is the set of structured node records associated with the nodes, and $\mathcal{R}\subseteq V$ is the set of designated result nodes corresponding to the final output entities. Each node $v_i$ stores an operation record
\[
\omega_i=(op_i,role_i,params_i,tags_i,context_i,\Delta_i),
\]
where $op_i$ specifies the canonical operation, $role_i$ identifies its functional role, $params_i$ stores parameterized inputs, $tags_i$ contains semantic annotations, $context_i$ records the modeling context, and $\Delta_i$ describes the induced semantic and topological changes. An edge $(v_i,v_j)$ exists when $v_j$ consumes an entity, parameter, constraint reference, or topology selection produced by $v_i$. Figure~\ref{fig:feature_graph_definition} illustrates operation dependencies, node records, and designated result nodes.

The construction graph serves as CADIR's serialized representation for cross-backend replay. CADIR lowers agent-facing composite calls into backend-independent canonical operations with parameter dependencies, while preserving sketch and assembly constraints for mapping to native backend operations.

Topology selection based on edge or face indices is brittle, as these indices may change after Boolean operations, parameter updates, or backend translation. To address this issue, we propose Geometric Signature Matching (GSM). CADIR represents each selected entity using a canonical geometric signature $\sigma$. Given a source signature $\sigma_s$ and a candidate signature $\sigma_c$, GSM computes their geometric difference as
\[
D(\sigma_s,\sigma_c)=\sum_{k\in\mathcal{K}} w_k E_k(\sigma_s,\sigma_c).
\]
Here, $\mathcal{K}$ includes bounding box, geometric type, center, length or area, endpoints, and surface normal; $E_k$ denotes the corresponding difference and $w_k\geq0$ its weight. Geometric errors are normalized by the model scale, and unavailable terms are omitted.

GSM first treats candidates satisfying $D(\sigma_s,\sigma_c)\leq\tau_e$ as exact matches, where $\tau_e<\tau_m$, and accepts the match only when the candidate is unique. Otherwise, it ranks candidates by $D$ and accepts the best candidate only when $d_1\leq\tau_m$ and $d_2-d_1\geq\delta$. Across CAD backends, one source edge may split into several target edges or several source edges may merge into one. GSM matches connected edge groups by curve support, endpoints, length, bounding box, and center, rejecting ambiguous mappings.

\begin{algorithm}[t]
\caption{Cross-Backend Geometric Signature Matching}
\label{alg:gsm}
\textbf{Input}: Source signature \(\sigma_s\), reconstructed target model \(B\)\\
\textbf{Output}: Matched target entity or failure
\begin{algorithmic}[1]
\STATE Canonicalize \(\sigma_s\) and collect target candidates \(\mathcal{C}\)
\FOR{each candidate \(c\in\mathcal{C}\)}
    \STATE Extract \(\sigma_c\) and compute \(d_c=D(\sigma_s,\sigma_c)\)
\ENDFOR
\STATE Let \(\mathcal{C}_e=\{c\in\mathcal{C}\mid d_c\leq\tau_e\}\)
\IF{\(|\mathcal{C}_e|=1\)}
    \STATE \textbf{return} the unique candidate in \(\mathcal{C}_e\)
\ENDIF
\STATE Let \(c_1,c_2\) have the two lowest scores \(d_1,d_2\)
\IF{\(d_1\leq\tau_m\) and \(d_2-d_1\geq\delta\)}
    \STATE \textbf{return} \(c_1\)
\ENDIF
\STATE Test split/merged edge mappings if unmatched
\STATE \textbf{return} the unique valid mapping, or failure
\end{algorithmic}
\end{algorithm}

We implement adapters for FreeCAD, SolidWorks, and Fusion~360. Each adapter traverses the construction graph in topological order and replays its parameters and constraints using native operations. When an operation references an edge or face, the adapter applies GSM to the current target geometry and passes the resolved entity or edge set to the native feature. This avoids backend-specific indices while preserving editable feature dependencies.

\subsection{Construction Graph Retrieval}

CADIR Agent uses construction graphs as structured indices for retrieval-augmented generation. Text and image queries retrieve relevant full graphs or subgraphs through two independently trained dual-tower models that use the same graph-encoder architecture but maintain separate parameters.

Each branch aligns its query modality with a branch-specific CAD-aware graph space. The text encoder $f_t(\cdot)$ maps a text query $q_t$ to $z_t=f_t(q_t)$, while the corresponding graph encoder $f_g^t(\cdot)$ maps a construction graph $G$ to $z_g^t=f_g^t(G)$. Similarly, the image encoder $f_i(\cdot)$ maps an image query $q_i$ to $z_i=f_i(q_i)$, while its graph encoder $f_g^i(\cdot)$ maps $G$ to $z_g^i=f_g^i(G)$. Their similarities are
\[
s_t(q_t,G)=z_t^\top z_g^t,\qquad
s_i(q_i,G)=z_i^\top z_g^i.
\]

In each branch, the graph encoder represents each node using its structured operation record $\omega_i$, while each directed edge represents an operation dependency. Gated message passing captures local operation dependencies, while a Graph Transformer models long-range relationships. The resulting node features are pooled into a graph-level representation. The text and image encoders are followed by separate projection layers to align each query modality with its corresponding graph space.

Both retrieval branches are trained independently using the same bidirectional contrastive objective. Each text or image query is paired with the full construction graph of its corresponding CAD model as a positive example. For either branch, given query and graph representation matrices $Z_q$ and $Z_g$, the contrastive objective is
\[
\mathcal{L}_{\mathrm{con}}
=\frac{1}{2}
\left(
\mathrm{CE}(S,P)+\mathrm{CE}(S^\top,P^\top)
\right),
\]
where $P$ is a normalized multi-positive correspondence matrix, and $S=Z_qZ_g^\top$ is the query--graph similarity matrix.

CADIR constructs a local subgraph by selecting a 3D feature operation as its local output and traversing backward along incoming dependency edges to collect the upstream nodes required for that feature. Given the resulting connected node subset $U\subseteq V$ that preserves the required dependencies, its local construction subgraph is
\[
G[U]=(U,E_U,\mathcal{O}_U,\mathcal{R}_U),
\]
where $E_U=\{(u,v)\in E\mid u,v\in U\}$, $\mathcal{O}_U=\{\omega_i\mid v_i\in U\}$, and $\mathcal{R}_U$ denotes its local output nodes. These local subgraphs receive no explicit supervision during training. Instead, they are encoded during offline indexing by the corresponding graph encoder trained on full-graph pairs, enabling zero-shot subgraph retrieval through the learned query--graph alignment.

Full construction graphs support whole-model matching, whereas subgraphs represent reusable modeling substructures such as sketch--extrusion--chamfer and base--rib--mounting-hole sequences.

\section{Experiments}

In this section, we evaluate CADIR in terms of intermediate representation, construction-graph retrieval, text-to-CAD generation, and cross-backend editability.

\subsection{Experimental Setup}

\subsubsection{Datasets.}
Unless stated otherwise, the generation experiments use 200 models, with 100 sampled from DeepCAD~\cite{wu2021Deepcad} and 100 from Fusion~360 Gallery~\cite{willis2021Fusion360}. The models are selected through stratified sampling across five command-count ranges: 1--10, 11--20, 21--30, 31--40, and 41--120, with 40 models sampled from each range. Each model is annotated with a natural-language description.

\subsubsection{Implementation details.}
All agent-based methods use GPT-5.4 as the base LLM. For text and image queries, we use Qwen3-Embedding-4B and gme-Qwen2-VL-2B-Instruct as the respective encoders. 

\subsubsection{Metrics.}
Following the evaluation protocol of CADDesigner~\cite{fan2026CADDesigner}, we assess generation quality using Intersection over Union (IoU), Chamfer Distance (CD), Hausdorff Distance (HD), and execution success rate (Suc.). Higher IoU and Suc. are better, whereas lower CD and HD are better. For retrieval, Recall@5 (R@5) and Recall@10 (R@10) indicate whether the target CAD model appears among the top 5 or top 10 results. For cross-backend evaluation, we report Node Reconstruction Rate (NRR), IoU, and Edit Success Rate (ESR). NRR measures the proportion of source construction-graph nodes successfully reconstructed in the target backend, IoU measures the geometric agreement of the reconstructed model, and ESR measures the proportion of editing tasks successfully completed after cross-backend reconstruction.

\subsection{Intermediate Representation Comparison}

We compare CQ~\cite{2026cadquery}, build123d~\cite{2025build123d}, HistCAD~\cite{dong2025HistCAD}, ForgeCAD~\cite{forgecad}, ECIP~\cite{fan2026CADDesigner}, and CADIR under the same agent framework and base LLM. For each representation, we provide only its official usage documentation as an agent skill, without examples or retrieval augmentation. As shown in Table~\ref{tab:ir_comparison}, CADIR achieves the best performance across all metrics, improving IoU by 4.9\% and reducing CD and HD by 8.2\% and 5.4\%, respectively, over the strongest baseline, while achieving an execution success rate of 1.0000.

\begin{table}[!ht]
\centering
\small
\renewcommand{\arraystretch}{1.05}
\begin{tabular*}{\columnwidth}{@{\extracolsep{\fill}}lcccc@{}}
\toprule
Representation & IoU $\uparrow$ & CD ($\times 10^{3}$) $\downarrow$ & HD $\downarrow$ & Suc. $\uparrow$ \\
\midrule
CQ & 0.2307 & 23.30 & 0.2580 & 0.9150 \\
build123d & 0.2577 & 19.99 & 0.2401 & 0.9850 \\
HistCAD & 0.2501 & 20.34 & 0.2405 & 0.8850 \\
ForgeCAD & 0.2382 & 20.62 & 0.2402 & 0.9300 \\
ECIP & 0.2490 & 21.87 & 0.2525 & 0.9650 \\
CADIR (Ours) & \textbf{0.2702} & \textbf{18.36} & \textbf{0.2272} & \textbf{1.0000} \\
\bottomrule
\end{tabular*}
\caption{Intermediate representation comparison results}
\label{tab:ir_comparison}
\end{table}

\subsection{Construction Graph Retrieval}

For retrieval experiments, we derive CADIR programs and construction graphs from 8,000 Text2CAD models, using 4,000 for training and the remaining 4,000 exclusively for evaluating retrieval recall. Neither retrieval split overlaps with the 200 models used for generation evaluation. We compare construction-graph indexing (CG) with direct CAD-program indexing (CP) using identical queries and candidate pools. As shown in Table~\ref{tab:retrieval_accuracy}, CG consistently outperforms CP for both text and image queries at both candidate-pool sizes. At $N=1000$, text-query CG exceeds CP by 0.4762 and 0.4008 in R@5 and R@10, respectively, indicating that explicit operation types and dependencies provide a more discriminative alignment target than raw program text.

\begin{table}[!ht]
\centering
\small
\setlength{\tabcolsep}{3pt}
\renewcommand{\arraystretch}{1.08}
\begin{tabular*}{\columnwidth}{@{\extracolsep{\fill}}llcccc@{}}
\toprule
Query
& Method
& \multicolumn{2}{c}{$N=200$}
& \multicolumn{2}{c}{$N=1000$} \\
\cmidrule(lr){3-4}
\cmidrule(lr){5-6}
&
& R@5 & R@10
& R@5 & R@10 \\
\midrule

\multirow{2}{*}{Image}
& CP
& 0.6150
& 0.7760
& 0.2778
& 0.4114 \\
& CG (Ours)
& \textbf{0.7826}
& \textbf{0.8852}
& \textbf{0.5409}
& \textbf{0.6513} \\
\midrule

\multirow{2}{*}{Text}
& CP
& 0.6078
& 0.7846
& 0.2733
& 0.4046 \\
& CG (Ours)
& \textbf{0.8650}
& \textbf{0.9106}
& \textbf{0.7495}
& \textbf{0.8054} \\
\bottomrule
\end{tabular*}
\caption{Retrieval accuracy averaged over 1,000 random candidate pools of size $N=200$ or $N=1000$. CG indexes construction graphs, whereas CP indexes CAD programs.}
\label{tab:retrieval_accuracy}
\end{table}

\begin{figure}[!h]
\centering
\includegraphics[width=0.9\columnwidth]{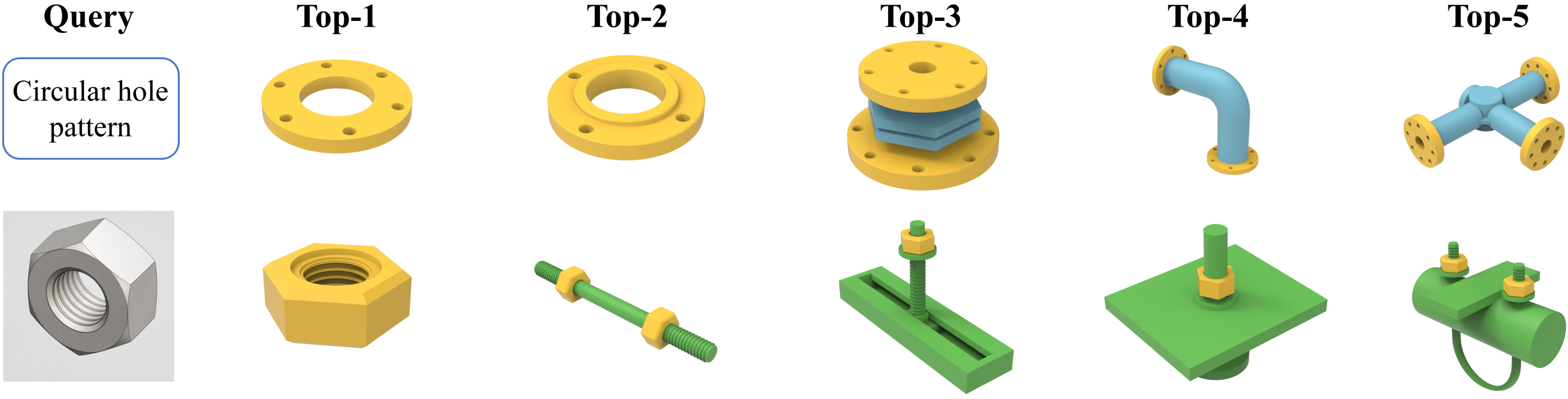}
\caption{Examples selected from text- and image-based retrieval results over 100 constructed candidate pools.}
\label{fig:retrieval_examples}
\end{figure}

Figure~\ref{fig:retrieval_examples} presents qualitative retrieval results for text and image queries. Both modalities recover models with similar global structures. For queries focused on screws, hole patterns, hexagonal nuts, or gears, subgraph retrieval also identifies models containing analogous operation compositions despite differences in overall shape. Construction graphs therefore support fine-grained construction reuse beyond whole-model matching.

\begin{table}[!ht]
\centering
\small
\renewcommand{\arraystretch}{1.08}
\begin{tabular*}{\columnwidth}{@{\extracolsep{\fill}}lcccc@{}}
\toprule
Setting & IoU $\uparrow$ & CD ($\times 10^{3}$) $\downarrow$ & HD $\downarrow$ & Suc. $\uparrow$ \\
\midrule

\multicolumn{5}{@{}l}{\textbf{Text Input}} \\
None
& 0.2702
& 18.36
& 0.2272
& 1.0000 \\
Full Graph
& 0.2837
& 17.03
& 0.2086
& 1.0000 \\
Full+Subgraph
& \textbf{0.3064}
& \textbf{13.47}
& \textbf{0.2007}
& 1.0000 \\
\midrule

\multicolumn{5}{@{}l}{\textbf{Image Input}} \\
None
& 0.2613
& 18.49
& 0.2263
& 1.0000 \\
Full Graph
& 0.3046
& 16.45
& 0.2011
& 1.0000 \\
Full+Subgraph
& \textbf{0.3852}
& \textbf{11.48}
& \textbf{0.1734}
& 1.0000 \\
\bottomrule
\end{tabular*}
\caption{Ablation of retrieval strategies on the 200-model generation set under text and image inputs.}
\label{tab:retrieval_ablation}
\end{table}

For generation-time retrieval, we construct a case library containing 100 validated CAD cases indexed by their construction graphs. These cases are drawn from a separate non-test split and do not overlap with the 200 models used for generation evaluation. As shown in Table~\ref{tab:retrieval_ablation}, full+subgraph retrieval consistently outperforms full-graph retrieval under both text and image inputs, with larger gains for image queries. This indicates that local subgraphs provide reusable modeling structures beyond whole-model matching and improve geometric fidelity.

\subsection{Text-to-CAD Generation}

\begin{table}[!bht]
\centering
\small
\renewcommand{\arraystretch}{1.05}
\begin{tabular*}{\columnwidth}{@{\extracolsep{\fill}}lcccc@{}}
\toprule
Method & IoU $\uparrow$ & CD ($\times 10^{3}$) $\downarrow$ & HD $\downarrow$ & Suc. $\uparrow$ \\
\midrule
Text2CAD
& 0.1578
& 56.54
& 0.3806
& 0.9500 \\
CADCodeVerify
& 0.2334
& 23.63
& 0.2566
& 0.8700 \\
CADDesigner
& 0.2347
& 23.87
& 0.2630
& 0.9950 \\
CADIR (Ours)
& \textbf{0.3064}
& \textbf{13.47}
& \textbf{0.2007}
& \textbf{1.0000} \\
\bottomrule
\end{tabular*}
\caption{Text-to-CAD generation results.}
\label{tab:text_to_cad}
\end{table}

\begin{table*}[!bht]
\centering
\renewcommand{\arraystretch}{1.12}
\begin{tabular*}{\textwidth}{
@{\extracolsep{\fill}}
lccccccccc
@{}
}
\toprule
& \multicolumn{3}{c}{FreeCAD}
& \multicolumn{3}{c}{Fusion 360}
& \multicolumn{3}{c}{SolidWorks} \\
\cmidrule(lr){2-4}
\cmidrule(lr){5-7}
\cmidrule(lr){8-10}

Method
& NRR $\uparrow$ & IoU $\uparrow$ & ESR $\uparrow$
& NRR $\uparrow$ & IoU $\uparrow$ & ESR $\uparrow$
& NRR $\uparrow$ & IoU $\uparrow$ & ESR $\uparrow$ \\
\midrule

Index-based
& 0.4490 & 0.7299 & 0.7148
& 0.4679 & 0.8187 & 0.7143
& 0.3175 & 0.6529 & 0.6190 \\

Point-sampling
& 0.5342 & 0.7986 & 0.8322
& 0.4173 & 0.8216 & 0.7177
& 0.3694 & 0.7029 & 0.6803 \\

GSM (Ours)
& \textbf{1.0000} & \textbf{0.9374} & \textbf{1.0000}
& \textbf{1.0000} & \textbf{0.9762} & \textbf{0.9796}
& \textbf{1.0000} & \textbf{0.9449} & \textbf{0.9184} \\
\bottomrule
\end{tabular*}

\caption{Comparison of cross-backend reconstruction and post-reconstruction editing.}
\label{tab:cross_backend_matching}
\end{table*}

We compare CADIR Agent with Text2CAD~\cite{khan2024Text2cad}, CADCodeVerify~\cite{alrashedy2025CADCodeVerify}, and CADDesigner~\cite{fan2026CADDesigner}. As shown in Table~\ref{tab:text_to_cad}, CADIR achieves the best results across all metrics, improving IoU by 30.5\% and reducing CD and HD by 43.0\% and 21.8\%, respectively, over the strongest baselines, while achieving an execution success rate of 1.0000.

\begin{figure}[!h]
\centering
\includegraphics[width=0.95\columnwidth]{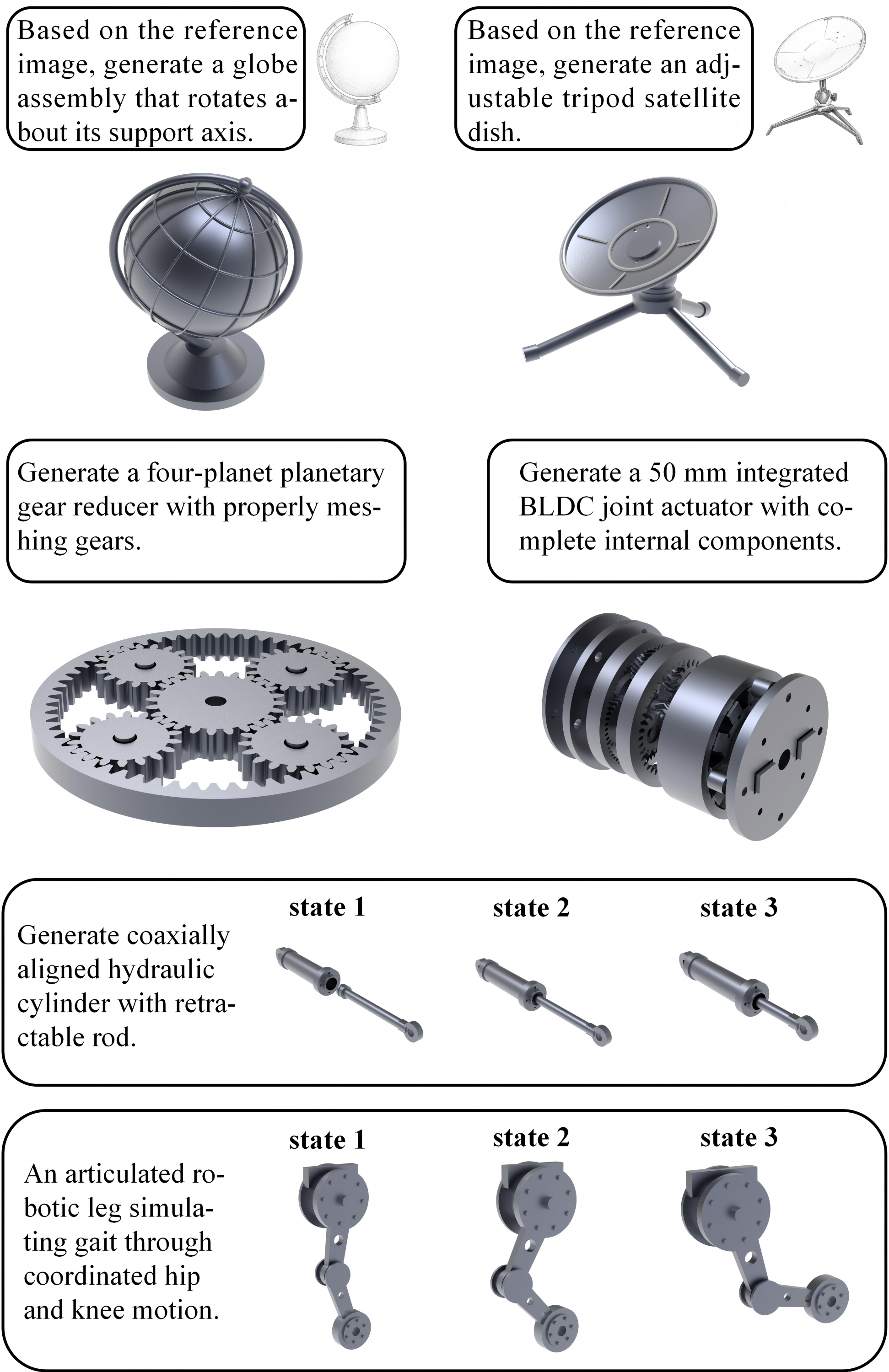}
\caption{CAD models generated by CADIR Agent.}
\label{fig:text_to_cad_examples}
\end{figure}
\FloatBarrier

Figure~\ref{fig:text_to_cad_examples} demonstrates CADIR's support for complex modeling operations, including mechanical assemblies, planetary gears, and integrated actuators. The last two examples illustrate assemblies with explicit constraints evaluated in multiple configurations. During generation, static collision checking helps detect component interference in each evaluated configuration.

\subsection{Cross-Backend Reconstruction and Editing}

\subsubsection{Cross-backend reconstruction.}
We use a development set to implement the backend adapters and tune the GSM weights and thresholds. The held-out test set contains 100 CADIR programs and construction graphs, covers all 115 CADIR modeling operations, and comprises 30,466 nodes. No test program is used for adapter development or hyperparameter selection. As shown in Table~\ref{tab:cross_backend_matching}, GSM achieves an NRR of 1.0000 in FreeCAD, Fusion~360, and SolidWorks. In comparison, the index-based baseline obtains NRR values of 0.4490, 0.4679, and 0.3175, while point-sampling~\cite{dong2025HistCAD} obtains 0.5342, 0.4173, and 0.3694, respectively. GSM also obtains IoU scores of 0.9374, 0.9762, and 0.9449, respectively. The remaining geometric differences arise from backend-specific geometry kernels, numerical tolerances, and native feature implementations.

\subsubsection{Post-reconstruction editing.}
Based on the same 100 held-out models, we construct 294 post-reconstruction editing tasks covering parameter, local-feature, and semantic-topology edits. An edit is counted as successful only if the target entity or parameter is uniquely resolved and correctly modified, the model recomputes without errors, and the intended change is reflected in the resulting geometry or feature state. As shown in Table~\ref{tab:cross_backend_matching}, GSM achieves ESR values of 1.0000, 0.9796, and 0.9184 in FreeCAD, Fusion~360, and SolidWorks, respectively, consistently outperforming both baselines. In Fusion~360 and SolidWorks, some reconstructed features lack equivalent native editable implementations or cannot be modified through the available interfaces, preventing ESR from reaching 1.0000.

\section{Conclusion}

We present CADIR, an executable intermediate representation for agentic CAD generation and cross-backend editing. By combining construction graphs, Geometric Signature Matching, and graph-based retrieval, CADIR enables reliable generation, reusable modeling structures, and native editable model reconstruction across CAD environments.

In future work, we will first incorporate simulation-based validation methods, such as kinematic simulation and dynamic analysis, to further improve the quality and engineering usability of generated models. We will also integrate domain knowledge to explore the deployment and application of CAD agents in real-world engineering scenarios.


\bibliography{refs}

\end{document}